\documentclass[AMA,STIX1COL]{WileyNJD-v2}
\usepackage{multirow}
\usepackage{booktabs}
\usepackage{adjustbox}
\usepackage{subcaption}

\newcommand{\M}[1]{\mbox {\textit{#1}}}

\articletype{Article Type}%

\received{26 April 2016}
\revised{6 June 2016}
\accepted{6 June 2016}

\begin{document}

\title{Text and author-level political inference using heterogeneous knowledge representations}

\author[1]{Samuel Caetano da Silva}

\author[2]{Ivandr\'e Paraboni*}

\authormark{da Silva and Paraboni}

\address[1]{\orgdiv{School of Arts Sciences and Humanities}, \orgname{University of S\~ao Paulo}, \orgaddress{\state{S\~ao Paulo}, \country{Brazil}}}

\address[2]{\orgdiv{School of Arts Sciences and Humanities}, \orgname{University of S\~ao Paulo}, \orgaddress{\state{S\~ao Paulo}, \country{Brazil}}}

\corres{*Ivandr\'e Paraboni\\\email{ivandre@usp.br}}

\presentaddress{Av Arlindo Bettio, 1000. S\~ao Paulo, Brazil 03828-000.}

\abstract[Summary]
{
The inference of politically-charged information from text data is a popular research topic in Natural Language Processing (NLP) at both text- and author-level. In recent years, studies of this kind have been  implemented with the aid of representations from transformers such as BERT. Despite considerable success, however, we may ask whether  results may be improved even further by combining  transformed-based models  with additional  knowledge representations. To shed light on this issue, the present work describes a series of experiments to compare alternative model configurations for political inference from text  in both English and Portuguese languages. Results suggest that certain text  representations - in particular, the combined use of BERT pre-trained language models with a syntactic dependency model - may outperform the alternatives across  multiple experimental settings, making a potentially strong case for further research in the use of heterogeneous text representations in these and possibly other NLP tasks.}



\maketitle

\footnotetext{\textbf{Abbreviations:} NLP, natural language processing; LSTM, long short-term memory networks; CNN, convolutional neural networks; BERT, Bidirectional Encoder Representations from Transformers, LIWC, Linguistic Inquiry and Word Count}

\section{Introduction}
\label{sec.intro}

Inferring politically-charged information from text data is a popular research topic in Natural Language Processing (NLP), with a wide range of applications to predict individuals' political views \cite{ranlp-wesley,bertha2019,bracis-pavan} and behaviour \cite{potthast2018,hp-semeval,li2021}.  Existing work in the field  often draws a distinction between two main tasks: the issue of political bias, and political ideology in general. Following \cite{hp-semeval} and others, political bias is presently defined as any extreme one-sided (or hyperpartisan)  discourse that, in a political context, clearly leans towards a liberal or conservative agenda, and which may be deliberately produced as a means to  convince or gather support. This contrasts with several other more nuanced forms of political expression, hereby called ideology, in which arguments tend to be more balanced, and which do no explicit promote a particular political agenda. These include expressions of political orientation (e.g., being left- or right-leaning) and political stance in general (e.g., being for or against a political party, a principle, or an individual), among  others. 

From a computational perspective, both  bias and ideology inference from text data may be further divided into two problem definitions, hereby called text- and author-level political inference. Text-level inference is more closely related to sentiment analysis \cite{deep-sentiment,sentrule,sentsoc} and stance classification \cite{st-post-sem, st-ensemble, bracis-pavan}, and it is intended to determine the meaning of an input text (e.g., whether the text expresses a liberal or conservative view.)  Less-known author-level inference, by contrast, is an instance of computational author profiling \cite{ca-demo, ca-ideology, ca-image, ca-bots,ca-fineg, ca-bert,ca-celeb}, that is, the task of inferring an individual's demographics (e.g., their political leaning)  based on samples of text that they have authored, and which may or may not convey politically-charged information explicitly.  

In recent years, both text- and author-level inference  have been  implemented with the aid of representations from transformers such as BERT \cite{bert} and others. Models of this kind - which may be seen as large, pre-trained language models - are able to capture deep contextual relations between words, and have been shown to significantly improve downstream task results \cite{lee2019,baly2020}. Despite the benefits afforded, however, in this paper we hypothesise that the use of pre-trained language models for political inference from text may be improved even further with the aid of additional text representations. More specifically, we ask how we may combine  transformed-based text representations  with  both syntactic information  and psycholinguistics-motivated features, and ask which of these so-called heterogeneous knowledge representations may have an impact on different  political inference tasks. 

To shed light on these issues, the present work takes the form of a series of experiments in supervised machine learning addressing a number of text- and author-level tasks alike. As in much of the existing work in the field, some of these  experiments will make use of English text data but, in addition to that, we will also address a number of tasks based on   Portuguese text data as well, and we will introduce a novel dataset  for this particular language. Our main contributions are summarised as follows.

\begin{itemize}
\item[i]{A neural architecture  that combines transformer-based language models, syntactic  dependencies and psycholinguistics-motivated features for political bias and ideology inference from text.}
\item[ii]{Text- and author-level formulations of the general political bias and ideology inference tasks.} 
\item[iii]{Experiments involving both mainstream English and less-studied Portuguese text data.}
\item[iv]{A novel, large dataset of  Twitter text data labelled with political stance  information.}
\end{itemize}

The reminder of this paper is structured as follows. Section \ref{sec.background} reviews existing work in political inference from text, and Section \ref{sec.tasks} describes the computational tasks to be addressed in this work. Section \ref{sec.method} presents our main approach to using heterogeneous knowledge representations. Section \ref{sec.results} summaries our experiment results, which are further discussed in Section \ref{sec.discuss}. Finally, Section \ref{sec.final} presents conclusions and future work.

\section{Related work}
\label{sec.background}

Table \ref{tab.related} presents an overview of recent NLP work on political bias and ideology inference from text data. All selected studies happen to be devoted to the English language. Further details are discussed below.

\begin{table}[htb!]
\begin{center}
\caption{Related work in political bias (top) and political ideology (bottom) inference from text. For each study, we report task category  as either `bias' (e.g., hyperpartisan news detection) or `ideology' (political ideology, stance, and alignment), label granularity  (t = text-level, a = author-level),  main learning features (text or network relations), learning methods (e.g., RNN = recurrent networks, U = stylometry, CNN = convolutional networks, LogReg = logistic regression, SVM = Support Vector Machines, RGC = graph networks, NPOV=neutral point of view), and text genre  (D = discourse/debate, N = news, T = Twitter, W = Wikipedia.)}

\label{tab.related}
\begin{tabular}{cccccc}
\hline
Study               & Category  & Labelling & Features   & Method      & Genre 
\\ \hline
\cite{potthast2018} & bias     &    t,a     & text        & U           & N\\
\cite{bertha2019}   & bias     &    t,a     & text        & CNN         & N\\
\cite{vernon2019}   & bias     &    t,a     & text        & LogReg      & N\\
\cite{drissi2019}   & bias     &    t,a     & text        & BERT        & N\\
\cite{lee2019}      & bias     &    t,a     & text        & BERT        & N\\
\cite{tintin2019}   & bias     &    t,a     & text        & SVM, LogReg & N\\
\cite{patankar2019} & bias     &     t      & text        & NPOV        & N\\
\cite{li2021}       & bias     &     t      & text        & biLSTM      & N\\
\hline
\cite{iyyer2014}    & ideology &     t      & text        & RNN         & D\\
\cite{bhatia2018}   & ideology &     t      & text        & LogReg      & N\\
\cite{kulkarni2018} & ideology &     t      & text,graph  & CNN         & N\\
\cite{baly2020}     & ideology &     t      & text        & LSTM, BERT  & N,T   \\
\cite{stefavov2020} & ideology &     t      & text,graph  & FastText    & T\\
\cite{feng2021}     & ideology &     a      & graph       & RGC         & W\\
\hline
\end{tabular}
\end{center}
\end{table}

\subsection{Political bias}

Extremely one-sided political bias has largely focused on the issue of  hyperpartisan news detection. The work in \cite{potthast2018}, which is among the first prominent NLP studies in this field, analyses a corpus of extremely one-sided news and presents a stylometry-based approach to distinguish  biased and neutral news, fake news and satire. Among other findings, results suggest that stylometry is of limited use in fake news detection, and that left- and right-wing news share a significant amount of stylistic similarities. 

Some of the results from \cite{potthast2018}  were taken as the basis for the influential SemEval-2019 shared task on hyperpartisan news detection in \cite{hp-semeval}. The  task comprised two formulations of the problem, each of them based on a different dataset. The {\M by\_article} dataset contained 1,273  manually labelled texts, and the {\M by\_publisher} dataset contained   754,000 articles labelled  via distant supervision (based on the publishing source of each article.) Results reported in \cite{hp-semeval} suggest that the {\M by\_publisher} task was generally  more challenging than the {\M by\_article} task. 

Among the participant systems at SemEval-2019, the work in \cite{bertha2019} reported the  overall highest accuracy in the {\M by\_article} task. The system makes use of  a pre-trained ELMo \cite{elmo} language model to encode news texts as input features to a CNN classifier. This approach will be taken as a baseline to our own experiments described in the next sections.

The work in \cite{vernon2019} makes use of a logistic regression classifier to detect hyperpartisan   news with the aid of hand crafted features (e.g., bias scores obtained from a bias lexicon, article- and sentence-level polarity,  subjectivity and modality scores etc.) and  Universal Sentence Encoder embeddings, among other alternatives. The system obtained the overall highest F1 score in the {\M by\_article} task at SemEval-2019. 

The studies  in \cite{drissi2019} and \cite{lee2019} are among the first to attempt using pre-trained BERT language models for hyperpartisan news detection. In both cases, however, results remain  below those obtained by several  more traditional approaches in both  {\M by\_article} and {\M by\_publisher} tasks.

The work in \cite{tintin2019} reported the  overall highest accuracy in the {\M by\_publisher} task using a Logistic Regression classifier with a bag-of-words text representation, outperforming a wide range of more complex models based on deep neural networks and others. 

The work in \cite{patankar2019} takes a more application-oriented approach to  hyperpartisan bias detection by presenting a  real-time system that flags political bias on news articles,  and then recommends similar articles from alternative sources. To this end, the system makes use of a bias lexicon  and unsupervised methods for clustering articles by topic similarity.

Finally, the work in \cite{li2021}, although not an original participant system at SemEval-2019, uses the SemEval {\M by\_article} dataset to address the hyperpartisan news detection task as well. The work makes use of a heterogeneous knowledge representation consisting of BERT pre-trained model enriched with social and political information and  linguistically-motivated features alike, and obtains results comparable to the top-performing systems at SemEval-2019 with the aid of a multihead Bi-LSTM architecture.

\subsection{Political ideology}

In addition to extremely one-sided bias, natural language text may convey many other more nuanced forms of political view, which are presently labelled as `ideology' for conciseness. The work in \cite{iyyer2014} is among the first of this kind, addressing the issue of political ideology (defined as left, right, or  neutral leaning)  detection in text. In what nowadays may be seen  as a standard approach to the task, the work uses  a recurrent neural network model and word embeddings built from  left- and right-leaning data to detect ideology in US congressman debates. Results suggest that this approach outperforms a range of logistic regression baseline systems using bag-of-words and word embeddings representations. 

Also in the US congressman debates domain, the work in \cite{bhatia2018} introduces a sentiment-oriented model to identify political ideology in text, the underlying assumption being that sentiment words may be revealing of an individual's political leaning (e.g., conservatives may arguably express more positive sentiment towards the free-market topic etc.) To investigate this issue, potentially relevant  topics were selected  from a debate corpus, including issues related to  health care, US military program and others, and topic-specific sentiments were computed as a probability distribution over ordinal polarity classes ranging from strongly positive to strongly negative. Results suggest that a logistic regression classifier based on  sentiment features outperforms the use  word embeddings and others. 

The work in \cite{kulkarni2018} addresses the issue of political ideology detection in news text using a so-called multi-view approach. In this approach, document-level features (e.g., the news headline and contents) and a network of links exhibited in the text are regarded as being complementary properties in the sense that authors may refer to links that  reinforce their political views. The study is carried out using a corpus of 120k  politically-related news in English, and results suggest that a  multi-view approach based on convolutional networks outperforms a range of baseline alternatives including the use of logistic regression classifiers and hierarchical attention models, among others.

The work in \cite{baly2020} addresses the task of detecting political ideology in  news texts from previously unseen media sources, which prevents models from learning the text source  rather than the ideology proper. The work makes use of complementary knowledge obtained from Twitter and Wikipedia sources, and uses adversarial adaptation and triplet loss pre-training (TLP) with both LSTM and BERT models. Results suggest that  combining TLP with additional Twitter information outperforms a range of alternatives, including the use of pre-trained transformer models alone, and those with access to additional Wikipedia information.

The work in \cite{stefavov2020} addresses the task of characterising the general political leaning of online media and influencers  by using  unsupervised learning  to determine the stance of Twitter users towards  a polarising topic based on their retweet behaviour, and then  performing label propagation to take  the resulting user stance information as training data for media political leaning detection.  Results suggest that a combination of  User-to-Hashtag and User-to-Mention graph embeddings with  BERT models built from both article titles and contents outperforms the use of these individual strategies in isolation. 

Finally, the work in \cite{feng2021} combines socially- and politically-related features to address the issue of entity stance prediction, an which  may be regarded as an instance of author-level inference as discussed in Section \ref{sec.intro}. Examples of entities under consideration include US presidents, parties, and states. The study builds a  heterogeneous information network  from Wikipedia articles, in which nodes are social entities and edges are  relations between them  (e.g., party affiliation, home state etc.) The network is combined with the Wikipedia text summary of each entity using a {\M RoBERTa} transformer  \cite{roberta} in a gated relational graph convolutional network for representation learning. Results suggest that the approach  outperforms a wide range of baseline alternatives, including bag-of-words, average word embeddings, transformers and graph-based models alike.

\section{Task definitions}
\label{sec.tasks}

The present work addresses two text-level political inference tasks (T1,T2) and three author-level tasks (T3,T4,T5) using data from three sources: the SemEval-2019 Hyperpartisan news corpus \cite{hp-semeval}, the BRmoral essay corpus \cite{brmoral} and a novel dataset, hereby called GovBR corpus, to be introduced in Section \ref{sec.corpus}. These  tasks are summarised in Table \ref{tab.tasks} and  discussed individually in the next sections. 

\begin{table}[hbt!]
\begin{center}
\caption{Text- and Author-level political inference tasks according to annotation level, target problem, class definition, corpus, dataset, and language.}
\label{tab.tasks}
\begin{tabular}{c c l l c c c}
\hline
Task & Level  & Target                & Classes                & Corpus   & Dataset       & Language\\
\hline
T1 & text   & hyperpartisan news    & hyperpartisan, neutral & SemEval  & by\_articles  & English\\
T2 & text   & political orientation & left, (centre), right  & BRmoral  & by\_opinion   & Portuguese\\
T3 & author & hyperpartisan news    & hyperpartisan, neutral & SemEval  & by\_publisher & English\\
T4 & author & political orientation & left, (centre), right  & BRmoral  & by\_author    & Portuguese\\
T5 & author & political stance      & for, against           & GovBR    &               & Portuguese\\
\hline
\end{tabular}
\end{center}
\end{table}

\subsubsection{Text-level tasks (T1,T2)}

Text-level tasks concern the inference of politically-charged information directly associated with the meanings of the input texts, in which case class labels are annotated  at the individual text level. As discussed in Section \ref{sec.intro}, this is analogous to sentiment analysis, stance classification and related tasks. 

Task T1 addresses the issue of text-level hyperpartisan news detection based on the SemEval {\M by\_articles} dataset \cite{hp-semeval}. This consists of a binary classification task intended to distinguish `hyperpartisan' from `neutral' information, as in the following examples. 

\begin{itemize}
\item Hyperpartisan: \textit{`Trump can't get Congress to repeal Obamacare, he's making changes that will penalize low-income people'}
\item Neutral: \textit{`Colin Kaepernick told a CBS reporter that he would represent the national anthem if he was hired by an NFL team'}
\end{itemize}

Task T2 addresses the issue of text-level political orientation  detection based on the BRmoral corpus \cite{brmoral} {\M by\_opinion} dataset following both binary (`left' / `right') and ternary (`left' / `centre' / `right') class definitions (to be further discussed in Section \ref{sec.brmoral}.) Examples taken from short essays related to the issue of same-sex marriage are illustrated as follows (translated from the original texts in Portuguese.) 

\begin{itemize}
\item Left-leaning: \textit{`Agreed, same-sex people must have their marriages valid, as they pay taxes and have the same civil obligations, so they should also have the same rights as the other.'}
\item Centre: \textit{`It's not up to the State to forbid this. However if a person asks me if I'm in favour, I'll say no, although I respect (their choice).'}
\item Right-leaning: \textit{`From the perspective of the civil institution, I do not see problems in marriage, however, the Christian doctrine makes it clear that the family consists of a man and a woman.'}
\end{itemize}

\subsubsection{Author-level tasks (T3,T4,T5)}
\label{sec.author.level}

Author-level tasks concern the inference of politically-charged information related to \textit{the individual} who wrote the input texts rather than to the literal meaning of the text. In this case, class labels are annotated at the author (or publisher) level. As discussed in Section \ref{sec.intro}, this is analogous to NLP author profiling. 

Task T3 is the author-level version of previous task T1, addressing the issue of author-level hyperpartisan news detection based on the SemEval {\M by\_publisher} dataset \cite{hp-semeval}. Once again, this consists of a binary classification task intended to distinguish `hyperpartisan' from  neutral' information, but using weakly labelled data determined by the source of information rather than  individually annotated texts. Thus, for instance, all texts produced by a publisher deemed to be a hyperpartisan source are labelled as `hyperpartisan' news regardless of their actual contents. 

Task T4 is the author-level version of previous task T2, addressing the issue of author-level political orientation  detection based on the BRmoral  \cite{brmoral} {\M by\_author} dataset following both binary (`left' / `right') and ternary (`left' / `centre' / `right') class definitions. To this end, all texts are weakly labelled with the self-reported political orientation of their authors regardless of the actual text contents. Thus, for instance, opinions that are annotated (at text-level) as `left-leaning' for the purpose of previous Task T2 may nevertheless be assigned a (author-level) `right-leaning' label if their authors happen to identify themselves as right-leaning individuals. 

Finally, task T5 addresses the issue of author-level stance classification based on the GovBR corpus to be described in the next section. This consists of a binary classification task intended to distinguish Twitter users who are `for' or `against' the current president of Brazil. In this case, tweets are weakly labelled with for/against stance information derived from popular hashtags. Thus, for instance, all tweets accompanied by a  {\M {\#RespectThePresident}} hashtag are assumed to be favourable to the president (the actual hashtags are not included in the data.) Examples of both classes are as follows. 

\begin{itemize}
\item For: \textit{`THIS IS MY PRESIDENT'}
\item Against: \textit{`This misgovernment is formed by indecent, immoral, ignorant, stupid and perverse people.'}
\end{itemize}

\section{Materials and method}
\label{sec.method}

The goal of the present study is to investigate the use of heterogeneous knowledge representations - based on  transformer-based language models, syntactic  dependencies, and psycholinguistics-motivated  features - for political bias and ideology inference from text as discussed in the previous section. In what follows we describe the data for each task, the classifier models to be investigated, and evaluation procedure.

\subsection{Data}
\label{sec.corpus}

The following sections discuss the three corpora to be taken as train and test data for our experiments, and present  descriptive statistics.

\subsubsection{SemEval-2019 hyperpartisan news corpus (tasks T1 and T3)}
\label{sec.semeval}
 
For the hyperpartisan news detection tasks T1 and T3, we will make use a subset of  the SemEval-2019 Hyperpartisan news corpus \cite{hp-semeval} in the English language. The SemEval-2019 corpus consists of political news organised in two datasets called {\M by\_articles} and {\M by\_publisher}, both of which annotated with `hyperpartisan' or `neutral' labels. Hyperpartisan news convey extreme one-sided information of either liberal or  conservative nature alike. The full corpus data - whose train subset is presently used in our experiments - originally conveys 645 news articles in the  {\M by\_articles} set, and 750 million articles in the {\M by\_publisher} set.

As discussed in the previous section, {\M by\_articles} texts are labelled individually according to their contents, and will be taken as an input to text-level hyperpartisan news detection (task T1). {\M by\_publisher} texts, by contrast, are weakly labelled according to their media source, and will be taken as the input for author-level hyperpartisan news detection (task T3). Despite using well-known shared task data, however, we notice that we do not presently seek to outperform the existing SemEval benchmark, but rather compare a number of novel computational strategies among themselves.

For the purpose of the present work, the original train portions of both datasets were  randomly split into development (80\%) and test (20\%) sets. Table \ref{tab.semeval} presents the resulting class distribution.
 
\setlength{\tabcolsep}{5pt} 
\begin{table}[htb!]
\centering
\caption{SemEval-2019 {by\_articles} and {by\_publisher} class distribution in development and test sets.}
\label{tab.semeval}
\begin{tabular}{l | ccc|ccc}
\hline
& \multicolumn{3}{c|}{{\M by\_articles}}           
& \multicolumn{3}{c}{{\M by\_publisher}}                \\
Set         & hyperpartisan & neutral       & Total & hyperpartisan     & neutral   & Total \\ \hline
Development & 332 {(64.3\%)}& 184 {(35.7\%)}& 516   & 1235 {(49.4\%)}   & 1265 {(50.6\%)} & 2500 \\
Test        & 75 {(58.1\%)} & 54 {(41.9\%)} & 129   & 492 {(49.2\%)}    & 508 {(50.8\%)}  & 1000 \\ 
\hline 
\end{tabular}
\end{table}

\subsubsection{BRmoral essay corpus (tasks T2 and T4)} 
\label{sec.brmoral}
 
For the political orientation detection tasks T2 and T4, we will make use of the BRmoral essay corpus \cite{brmoral} in the Portuguese language. This consists of  short essays about eight topics of liberal and conservative nature alike (same-sex marriage, gun possession, abortion, death penalty, drug legalisation, lowering of criminal age, racial quotas, and tax exemptions for churches) labelled with both stance scores (from `totally against' to `totally for' each topic) and authors' demographics, including their self-reported political orientation from `extreme left' to `extreme right'. The full corpus data conveys 4080 essays  written by 510 crowd-sourced volunteers \cite{brmoral}.

The dual labelling scheme in the BRmoral corpus  (i.e., either based on individual stances or author's own  political orientation)  gives rise to two dataset definitions, hereby called {\M by\_opinion} and {\M by\_author} for analogy with the SemEval {\M by\_articles} and {\M by\_publisher} datasets discussed in the previous section. Both {\M by\_opinion} and {\M by\_author}  are labelled with `left', `right' and, depending on the task under consideration (see below), also with `centre' information. 

{\M by\_opinion} takes as labels the liberal and conservative stance information available from the corpus to determine, albeit indirectly, a text's likely political leaning. More specifically, texts expressing an opinion against so-called liberal topics (same-sex marriage, abortion, drug legalisation, and racial quotas), or those expressing opinions in favour of so-called conservative topics (death penalty, gun possession, lowering of criminal age, and tax exemptions for churches), are labelled as `right', and so forth. This dataset will be taken as an input to text-level political orientation  detection (task T2).

The {\M by\_author} dataset, by contrast, takes as labels the actual authors' political orientation information available from the corpus, in what may be seen as an instance of  weakly labelling not unlike the SemEval {\M by\_publisher} labels discussed in the previous section. BRmoral {\M by\_author}  texts will be taken as an input to author-level political orientation  detection (task T4). 

Both datasets were  randomly split into development (80\%) and test (20\%) sets. Table \ref{tab.brmoral} presents the resulting class distribution.

\setlength{\tabcolsep}{5pt} 
\begin{table}[htb!]
\centering
\caption{BRmoral {by\_opinion} and {by\_author} class distribution in development and test sets.}
\label{tab.brmoral}
\begin{tabular}{l | cccc | cccc}
\hline
&\multicolumn{4}{c|}{\M by\_opinion}                    
& \multicolumn{4}{c}{\M by\_author} \\
Set         & left    & centre    & right & Total & left    & centre     & right & Total \\ 
\hline 
Development & 1201 (36.8\%) & 685 (21.0\%) & 1378 (42.2\%)  & 3264  & 1210 (37.1\%) & 1158 (35.5\%) & 896 (27.4\%)   & 3264  \\
Test        & 299 (36.6\%)  & 176 (21.6\%) & 341 (41.8\%)   & 816   & 310 (38.0\%)    & 282 (34.5\%)  & 224 (27.5\%)   & 816   \\ 
\hline
\end{tabular}
\end{table}

\subsubsection{GovBR political stance corpus (task T5)}
\label{sec.govbr}
 
Finally, for the political stance task T5, we created a novel language resource based on Twitter data in the Portuguese language, hereby called the GovBR corpus. GovBR comprises  a collection of tweets written by users who expressed a clear stance towards the current president of Brazil. The corpus was built by selecting two disjoint sets of users - supporters and opponents of the said president  - according to  the use of a number of popular politically-charged hashtags (e.g., {\M {\#RespectThePresident}} or {\M {\#NotHim}}. The full corpus data - from which non-political tweets were filtered out as discussed below - conveys 13.5 million tweets written by 5452 unique users.

For each selected user, all their publicly available tweets (i.e., disregarding their retweets) were downloaded. Users who simultaneously promoted supportive and opposing hashtags were discarded, and so were all hashtags and all messages shorter than five words. Finally, tweets that did not convey a minimal level of political content were also discarded. To this end, we computed a TF-IDF representation of the political section of the \textit{Folha de SP} newspaper\footnote{\url{https://www.kaggle.com/marlesson/news-of-the-site-folhauol}}, and kept only the tweets conveying a minimum  degree of similarity to the political news texts. After the removal of non-political tweets, we obtained approximately 25 tweets per user.

GovBR politically-related tweets will be taken as the input for author-level political stance detection (task T5). In our current work, we use a balanced subset of this data  consisting of 4000 randomly selected tweets. These  were randomly split into development (80\%) and test (20\%) sets as illustrated in Table \ref{tab.govbr}

 \setlength{\tabcolsep}{5pt} 
\begin{table}[htb!]
\centering
\caption{GovBR class distribution in development and test sets.}
\label{tab.govbr}
\begin{tabular}{l ccc}
\hline
Set         &  against & for & Total \\ 
\hline
Development & 1600 (49.9\%) & 1608 (50.1\%) & 3208  \\
Test        & 405 (50.5\%)  & 397 (49.5\%)  & 802   \\ 
\hline
\end{tabular}
\end{table}

\subsection{Models}

As a means to investigate the issue of political inference from text (tasks T1-T5 described in the previous sections), in what follows we propose combining heterogeneous knowledge representations into a convolutional neural network architecture. An overview of this architecture is presented in Section \ref{sec.architecture}, and its individual components are described in Section \ref{sec.components}.

\subsubsection{Architecture}
\label{sec.architecture}

\begin{figure}[t]
\centerline{\includegraphics[width=.9\textwidth]{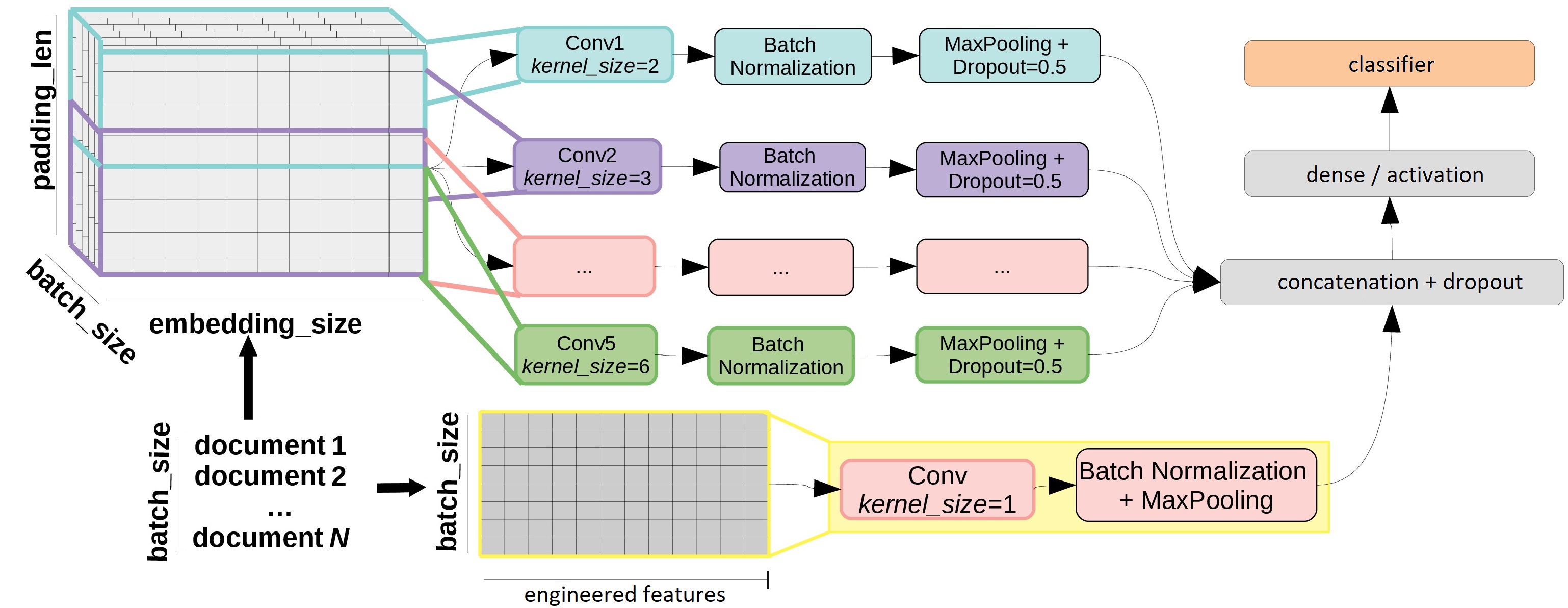}}
\caption{General architecture.\label{fig.pipeline}}
\end{figure}

We envisaged a convolutional neural network for political inference from text that combines three kinds of text representation: (i) pre-trained language models provided by Bidirectional Encoder Representations from Transformers (BERT) \cite{bert}, hereby called {\it bert}; (ii) syntactic bigram counts computed from  dependency graphs \cite{sngram}, hereby called {\it sngram}, and (iii) psycholinguistics-motivated features obtained from Linguistic Inquiry and Word Count (LIWC) \cite{liwc} and from the Medical Research Council (MRC) database \cite{mrc}, hereby called {\it psych}. This  architecture is illustrated in Figure \ref{fig.pipeline} and further discussed below.
 
Given a set of input documents (bottom left of the figure), we use a text classifier model that takes as an input both  standard text features represented as contextual embeddings  {\it bert} (top left), and engineered features (bottom centre) that combine the alternative text representations based on syntactic dependencies {\it sngram} and psycholinguistics-motivated features {\it psych}. In this approach, {\it bert} embeddings are taken as an input to five convolutional layers (Conv1-Conv5) followed by batch normalisation, a max pooling and a 0.5 dropout layers, whereas engineered features are processed in parallel by a kernel of size 1, also  followed by a normalisation layer  and max pooling. Finally, the output of the operations Conv1-Conv5 is concatenated with the engineered feature vector followed by a 0.5 dropout, and a softmax activation function produces the class predictions.

The full model architecture, hereby called  {\M bert+sngram+psych}, will be evaluated against a number of baseline systems, and also against some of its individual components as discussed in Section \ref{sec.procedure}.

\subsubsection{Text representations}
\label{sec.components}

In the {\M bert+sngram+psych} model, input  texts are to be represented as contextual embeddings (our {\it bert} component), syntactic bigram counts {\it sngram}, and psycholinguistics-motivated features ({\it psych}). These components are described  as follows. 

Bidirectional Encoder Representations from Transformers (BERT) \cite{bert} are now  mainstream in NLP and related fields, and are the basis for our {\it bert} architectural component as well. This consists of pre-trained BERT language models  fine-tuned for  each of our classification tasks T1-T5. For the English language models (tasks T1 and T3), we use  \textit{base-uncased} BERT, and for the Portuguese language (tasks T2, T4, and T5) we use  \textit{multilingual-base-uncased} BERT.
 
The \textit{sngram} component explores the use of text structural information  to enrich our classifier models by  computing  syntactic bigram counts from dependency graphs. To this end, we first compute a syntactic dependency graph from the input text using SpaCy\footnote{\url{https://spacy.io/usage/linguistic-features\#dependency-parse}}, and then generate a TF-IDF bigram model  as suggested in \cite{sngram}.  For the English language models (tasks T1 and T3), we use the {en\_core\_web\_sm} pipeline, and for the Portuguese language (tasks T2, T4, and T5) we use  the {pt\_core\_news\_sm} pipeline. Finally, we perform univariate feature selection using F1 as a score function in order to keep only the  $k$ best features. Optimal $k$ values for each task were obtained through grid search on development data.

The \textit{psych} component makes use of psycholinguistics-motivated features computed with the aid of both LIWC \cite{liwc} and MRC \cite{mrc} lexicons. Examples of LIWC  word categories include those related to attention focus (e.g., pronouns and verb tense), affective or emotional processes (positive and negative emotions, anxiety, fear etc.), social relationships (e.g., family, friends etc.) and others. Similarly, MRC categories cover lexical features such as concreteness, age of acquisition, and others. For the English language models (tasks T1 and T3), we use the 93-feature LIWC-2015 lexicon \cite{liwc2015}  and the 9-feature MRC database \cite{mrc}. For  the Portuguese language (tasks T2, T4, and T5) we use 64-feature LIWC-BR \cite{liwc-br} and the 6 MRC-like features from \cite{PsychoProps}.  

Both LIWC and MRC text representations consist of word category counts normalised by document size, in which words that belongs to more than one category update all related counts (e.g., `she' is a pronoun and also a feminine word etc.) Both representations are concatenated as a single vector of size $93 + 9 = 101$ features for English, or $64 + 6 = 70$ features for Portuguese. As in the case of the syntactic features discussed above, we once again perform univariate feature selection with F1 as a score function to obtain the $k$ best features for each task.

\subsection{Procedure}
\label{sec.procedure}

We conducted a series of experiments focused on tasks T1-T5 introduced in Section \ref{sec.tasks}  to assess the use of the {\M bert+sngram+psych} model and some of its subcomponents, namely, {\it bert}, {\it sngram}, {\it psych} alone, and also the two BERT-based   pairs   {\M bert+sngram} and {\M bert+psych}. More specifically, our goal is to  investigate how each of these alternatives  compare to two baseline systems, namely, the Bertha von Suttner model described in \cite{bertha2019}, which was the overall best-performing system at SemEval-2019 Hyperpartisan news detection shared task \cite{hp-semeval}, and the use of BERT alone as a classifier, hereby called {\M BERT.baseline}.

All models were trained in 30 epochs using a development dataset partition, and then evaluated using a previously unseen test data as described in Section \ref{sec.corpus}. For BERT-based models, additional pre-processing was performed to remove all non-alphabetic characters, links and HTML tags. All input documents were limited to their first 300 tokens, and shorter  documents were completed with the [PAD] token. This representation was taken as the input to a BERT model of size 768, resulting in text embeddings of size  $300  \times 768$.
  
{\M sngram} and {\M psych} features are concatenated as a single vector, and a z-score function is applied to obtain standardised value ranges.  Table \ref{features-psico-sngram} summarises the actual number of features considered by each models and for each corpus.
 
\begin{table}[htb!]
\centering
\caption{Number of features used in \textit{psych} and \textit{sngram} vectors in each corpus.}
\label{features-psico-sngram}
\begin{tabular}{ccc}
\hline
Corpus                      & {psych}   & {sngram}  \\ 
\hline
{\M SemEval by\_articles}   & 37        & 13,107     \\
{\M SemEval by\_publisher}  & 78        & 38,220     \\
{\M BRmoral by\_opinion}    & 55        & 13,466     \\
{\M BRmoral by\_author}     & 22        & 13,399     \\ 
{GovBR}                     & 25        &  6,998     \\
\hline
\end{tabular}
\end{table}

Evaluation proper was carried out by (i) measuring  Accuracy (Acc), macro $F_1$  ($F_1$), Precision (P) and Recall (R) scores, (ii) by assessing statistical significant differences between models, and (iii) by providing model prediction explanations. To this end, statistical significance is assessed by using the McNemar test \cite{mcnemar} in the case of  binary classifiers, and by using the Stuart-Maxwell test \cite{stuart1955test, maxwell1970comparing} for ternary classifiers. For each task, two kinds of significance tests are conducted. First, we identify those models that are statistically superior to the reference model in \cite{bertha2019}. Second, we  identify the groups of models that are statistically distinguishable from others. Finally, we performed eli5 prediction explanation\footnote{\url{https://eli5.readthedocs.io/en/latest/}} to compute the word features more strongly correlated with each class and task, as discussed in Section \ref{sec.discuss}.

\section{Results}
\label{sec.results}  

This section presents results of the full model {\M bert+sngram+psych} and its subcomponents ({\it bert}, {\it sngram}, {\it psych}, {\M bert+sngram}, and {\M bert+psych}) compared against  those obtained by the work  in  \cite{bertha2019} and by  BERT alone as a classifier ({\M bert.baseline}). 

Results are to be reported as Accuracy (Acc), $F_1$, Precision (P) and Recall (R) scores divided into three groups: (1) on the top row baselines from \cite{bertha2019} and {\M bert.baseline}, (2) following by model components \textit{bert}, \textit{sngram}, \textit{psych}, {\M bert+sngram}, and {\M bert+psych},  and (3) on the bottom row the full model {\M bert+sngram+psych}. In all scenarios, best accuracy scores are highlighted, and also marked as * when found to be statistically superior to the baseline system in \cite{bertha2019}. Finally, in addition to the main results report for each task, models are also depicted in statistically significant clusters ($p < 0.05$) according to their accuracy scores.

\subsection{Task T1: Text-level hyperpartisan news detection }
\label{sec.t1}

Results for task T1 - text-level hyperpartisan news detection - using the SemEval corpus {\M by\_articles} dataset are summarised in Table \ref{tab.t1.results} and further discussed below.

\begin{table}[hbt!]
\centering
\caption{Text-level hyperpartisan news detection results. }
\label{tab.t1.results}
\begin{tabular}{l cccc }
\hline
Model             & Acc             & $F_1$   & P    & R  \\ 
\hline
Jiang et. al.             & 0.72            & 0.65 & 0.69 & 0.61\\
BERT.baseline     & 0.68            & 0.65 & 0.61 & 0.71\\
                  &                 &      &      &     \\
bert              & 0.75            & 0.69 & 0.73 & 0.65\\
sngram            & 0.69            & 0.69 & 0.70 & 0.69\\
psych             & 0.47            & 0.46 & 0.45 & 0.47\\
bert+sngram       & \textbf{0.78*}  & 0.77 & 0.78 & 0.78\\
bert+psych        & 0.74            & 0.71 & 0.66 & 0.76\\
                  &                 &      &      &     \\
bert+sngram+psych & 0.67            & 0.64 & 0.67 & 0.67\\ 
\hline
\end{tabular}
\end{table}

Based on these results, we notice that the best-performing model is {\M bert+sngram}. The difference between this and the  baseline in \cite{bertha2019} is statistically significant ($\chi = 5.161$, $\alpha$ = 0.05, $p$ < 0.05). The full model {\M bert+sngram+psych}, by contrast, ranks considerably lower. To further illustrate this outcome, the models were clustered into homogeneous groups (A,B,C) by statistical significance according to their accuracy scores as illustrated in Table \ref{tab.t1.homo}.
 
\begin{table}[htb!]
\centering
\caption{Text-level hyperpartisan news detection homogeneous groups. }
\label{tab.t1.homo}
\begin{tabular}{l c ccc}
\hline
Model               & Acc & \multicolumn{3}{c}{Groups} \\ 
\hline
bert+sngram         & \textbf{0.78*}    & A &   & \\
bert                & 0.75              & A &   & \\
bert+psych          & 0.74              & A &   & \\
Jiang et al.        & 0.72              &   & B & \\
sngram              & 0.69              &   & B & \\
\mbox{BERT.baseline}& 0.68              &   & B  & \\
bert+sngram+psych   & 0.67              &   & B  & \\
psych               & 0.47              &   &    & C \\ 
\hline
\end{tabular}
\end{table}

We notice that, in addition to the best-performing {\M bert+sngram} model, both \textit{bert} and {\M bert+psych} obtain, to a lesser extent, statistically similar results within group A. The full model and the reference baseline in \cite{bertha2019}, by contrast, are both members of group B.





\subsection{Task T2: Text-level political orientation detection}
\label{sec.t2}
 
Results for task T2 - text-level political orientation detection - using the BRmoral corpus {\M by\_opinion} dataset for both binary (left, right) and ternary (left,centre,right) classification are summarised in Table \ref{tab.t2.results} and further discussed below.

\begin{table}[htb!]
\centering
\caption{Text-level political orientation detection results.}
\label{tab.t2.results}
\begin{tabular}{l cccc | cccc}
\hline
\multicolumn{1}{c}{}&
\multicolumn{4}{c|}{Binary classification}&
\multicolumn{4}{c}{Ternary classification}\\
Model               & Acc           & $F_1$    & P    & R    & Acc            & $F_1$   & P    & R\\ 
\hline
Jiang et. al.       & 0.76          & 0.76  & 0.77 & 0.76  & 0.62          & 0.61 & 0.61 & 0.62\\
\mbox{BERT.baseline}& 0.57          & 0.57  & 0.57 & 0.57  & 0.43          & 0.36 & 0.43 & 0.43\\
&&&&&&&&\\
bert                & 0.75          & 0.75  & 0.76 & 0.75  & 0.60          & 0.56 & 0.59 & 0.60\\
sngram              & 0.70          & 0.70  & 0.71 & 0.70  & 0.55          & 0.53 & 0.55 & 0.55\\
psych               & 0.68          & 0.68  & 0.68 & 0.68  & 0.52          & 0.50 & 0.49 & 0.52\\
bert+sngram         & \textbf{0.78} & 0.78  & 0.78 & 0.78  & \textbf{0.64} & 0.62 & 0.63 & 0.64\\
bert+psych          & 0.76          & 0.76  & 0.77 & 0.76  & 0.60          & 0.59 & 0.59 & 0.60\\
&&&&&&&&\\
bert+sngram+psych   & 0.73          & 0.73  & 0.74 & 0.73  & 0.58          & 0.58 & 0.58 & 0.58\\ 
\hline
\end{tabular}
\end{table}

Once again, the best-performing model for both binary and ternary classification is {\M bert+sngram}. However, differences between this  and others, including the baseline in \cite{bertha2019}, were not found to be statistically significant. To further illustrate this outcome, homogeneous groups related to the binary classification task are shown in Table \ref{tab.t2.bin.homo}, and groups related to ternary classification are shown  in Table \ref{tab.t2.ter.homo}.  
 
\begin{table}[htb!]
\centering
\caption{Binary text-level political orientation detection homogeneous groups. }
\label{tab.t2.bin.homo}
\begin{tabular}{l c cccc}
\hline
Model               & Acc & \multicolumn{3}{c}{Groups} \\ 
\hline
bert+sngram         & \textbf{0.78} & A &   &   & \\
Jiang et. al.       & 0.76          & A &   &   & \\
bert+psych          & 0.76          & A &   &   & \\
bert                & 0.75          & A &   &   & \\
bert+sngram+psych   & 0.73          &   & B &   & \\
sngram              & 0.70          &   & B &   & \\
psych               & 0.68          &   &   & C & \\
\mbox{BERT.baseline}& 0.57          &   &   &   & D \\ 
\hline  
\end{tabular}
\end{table}

\begin{table}[htb!]
\centering
\caption{Ternary text-level political orientation detection homogeneous groups. }
\label{tab.t2.ter.homo}
\begin{tabular}{l c cccc}
\hline
Model               & Acc & \multicolumn{3}{c}{Groups} \\ 
\hline
bert+sngram         & \textbf{0.64} & A     &   &   &   \\
Jiang et. al.       & 0.62          & A     &   &   &   \\
bert                & 0.60          & A     &   &   &   \\
bert+psych          & 0.60          &       & B &   &   \\
bert+sngram+psych   & 0.58          &       & B &   &   \\
sngram              & 0.55          &       & B &   &   \\
psych               & 0.52          &       &   & C &   \\
\mbox{BERT.baseline}& 0.43          &       &   &   & D \\ 
\hline
\end{tabular}
\end{table}

In both binary and ternary classification tasks, although {\M bert+sngram} still obtains the highest accuracy scores, its  simpler \textit{bert} sub-component  (i.e., the combination of a fine-tuned BERT  model with a CNN classifier, as discussed in Section \ref{sec.method}) is statistically similar. The full model, {\M bert+sngram+psych}, composes the group B in both classification tasks, ranking lower than the reference system \cite{bertha2019}.



\subsection{Task T3: Author-level hyperpartisan news detection }
\label{sec.t3}

Results for task T3 - author-level hyperpartisan news detection - using the SemEval corpus {\M by\_publisher} dataset are summarised in Table \ref{tab.t3.results} and further discussed below.

\begin{table}[htb!]
\centering
\caption{Author-level hyperpartisan news detection results. }
\label{tab.t3.results}
\begin{tabular}{l cccc }
\hline
Model               & Acc             & $F_1$   & P    & R  \\ 
\hline
Jiang et. al.       & 0.56            & 0.62 & 0.55 & 0.71 \\
\mbox{BERT.baseline}& 0.54            & 0.60 & 0.51 & 0.75 \\
                    &                 &      &      &      \\
bert                & 0.58            & 0.67 & 0.55 & 0.85 \\
sngram              & 0.57            & 0.55 & 0.57 & 0.57 \\
psych               & 0.53            & 0.53 & 0.53 & 0.53 \\
bert+sngram         & 0.57            & 0.54 & 0.59 & 0.57 \\ 
bert+psych          & \textbf{0.61*}  & 0.67 & 0.58 & 0.80 \\
                    &                 &      &      &      \\
bert+sngram+psych   & 0.55            & 0.52 & 0.50 & 0.55 \\ 
\hline
\end{tabular}
\end{table}

Based on these results, we notice that the best-performing alternative is {\M bert+psych}. The difference between this and the baseline in \cite{bertha2019} is statistically significant ($\chi$ = 6.715, $\alpha$ = 0.05, $p$ < 0.01). Models were clustered into homogeneous groups by statistical significance according to their accuracy scores as illustrated in Table \ref{tab.t3.homo}.
 
\begin{table}[htb!]
\centering
\caption{Author-level hyperpartisan news detection homogeneous groups. }
\label{tab.t3.homo}
\begin{tabular}{l c ccc}
\hline
Model               & Acc & \multicolumn{3}{c}{Groups} \\ 
\hline
bert+psych          & \textbf{0.61*}  & A &   \\
bert                & 0.58            & A &   \\
bert+sngram         & 0.57            & A &   \\
sngram              & 0.57            &   & B \\
Jiang et al.        & 0.56            &   & B \\
bert+sngram+psych   & 0.55            &   & B \\
\mbox{BERT.baseline}& 0.54            &   & B \\
psych               & 0.53            &   & B \\ 
\hline
\end{tabular}
\end{table}

According to Table \ref{tab.t3.homo}, group A includes most alternatives  that  are based on BERT language models, and once again the difference between the full model  {\M bert+sngram+psych} and the reference baseline in \cite{bertha2019}, both of which in group B, is not statistically significant.

 

\subsection{Task T4: Author-level political orientation detection}
\label{sec.t4}

Results for task T4 - author-level political orientation detection - using the BRmoral corpus {\M by\_author} dataset for both binary (left, right) and ternary (left,centre,right) classification are summarised in Table \ref{tab.t4.results} and further discussed below.

\begin{table}[htb!]
\centering
\caption{Author-level political orientation detection results.}
\label{tab.t4.results}
\begin{tabular}{l cccc | cccc}
\hline
\multicolumn{1}{c}{}&
\multicolumn{4}{c|}{Binary classification}&
\multicolumn{4}{c}{Ternary classification}\\
Model             & Acc           & $F_1$   & P    & R    & Acc            & $F_1$   & P    & R\\ 
\hline
Jiang et al.      & 0.61          & 0.61 & 0.61 & 0.61 & 0.38          & 0.37 & 0.37 & 0.38\\
\mbox{BERT.baseline} & 0.56          & 0.55 & 0.55 & 0.56 & 0.37          & 0.31 & 0.37 & 0.37\\
                  &                 &      &      &      &&&&\\
bert              & 0.53          & 0.53 & 0.58 & 0.53 & 0.41          & 0.37 & 0.41 & 0.41\\
sngram            & 0.56          & 0.56 & 0.57 & 0.56 & 0.38          & 0.37 & 0.37 & 0.38\\
psych             & 0.59          & 0.56 & 0.57 & 0.59 & 0.37          & 0.35 & 0.35 & 0.37\\
bert+sngram       & \textbf{0.63} & 0.63 & 0.63 & 0.63 & 0.41          & 0.37 & 0.40 & 0.41\\
bert+psych        & 0.60          & 0.53 & 0.59 & 0.60 & 0.36          & 0.30 & 0.44 & 0.36\\
                  &                 &      &      &      &&&&\\
bert+sngram+psych & 0.59          & 0.59 & 0.61 & 0.59 & \textbf{0.42} & 0.41 & 0.41 & 0.42\\ 
\hline
\end{tabular}
\end{table}

The best-performing model for binary classification  is {\M bert+sngram}, and for ternary classification is {\M bert+sngram+psych}. However, the differences between these and the baseline in \cite{bertha2019} were not found to be significant. Homogeneous groups for the binary task  are illustrated in Table \ref{tab.t4.bin.homo}, and groups for the  ternary task  are illustrated in Table \ref{tab.t4.ter.homo}.  
 
\begin{table}[htb!]
\centering
\caption{Binary author-level political orientation detection homogeneous groups. }
\label{tab.t4.bin.homo}
\begin{tabular}{l c cccc}
\hline
Model               & Acc & \multicolumn{3}{c}{Groups} \\ 
\hline
bert+sngram         & \textbf{0.63} & A &   \\
Jiang et al.        & 0.61          & A &   \\
bert+psych          & 0.60          & A &   \\
psych               & 0.59          & A &   \\
bert+sngram+psych   & 0.59          & A &   \\
\mbox{BERT.baseline}& 0.56          & A &   \\
sngram              & 0.56          &   & B \\
bert                & 0.53          &   & B \\ 
\hline
\end{tabular}
\end{table}

\begin{table}[htb!]
\centering
\caption{Ternary author-level political orientation detection homogeneous groups. }
\label{tab.t4.ter.homo}
\begin{tabular}{l c cccc}
\hline
Model               & Acc & \multicolumn{3}{c}{Groups} \\ 
\hline
bert+sngram+psych   & \textbf{0.42}   & A &       \\
bert+sngram         & 0.41            & A &       \\
bert                & 0.41            & A &       \\
Jiang et al.        & 0.38            & A &       \\
sngram              & 0.38            & A &       \\
psych               & 0.37            &   & B     \\
\mbox{BERT.baseline}& 0.37            &   & B     \\
bert+psych          & 0.36            &   & B     \\ 
\hline
\end{tabular}
\end{table}

In both binary and ternary classification tasks, we notice that several models turned out to obtain  statistically equivalent results. This outcome, which is similar to what has been  observed in  the text-level political orientation task (cf. Section \ref{sec.t2}) based on the same corpus, will be further discussed in Section \ref{sec.discuss}.

  

\subsection{Task T5: Author-level political stance detection}
\label{sec.t5}

Results for task T5 - author-level political stance detection -  using the {\M GovBR} corpus are summarised in Table \ref{tab.t5.results} and further discussed below.

\begin{table}[htb!]
\centering
\caption{Author-level political stance detection results. }
\label{tab.t5.results}
\begin{tabular}{l cccc }
\hline
Model             & Acc           & $F_1$   & P    & R    \\ 
\hline
Jiang et. al.     & 0.58          & 0.55 & 0.62 & 0.58 \\
\mbox{BERT.baseline} & 0.53          & 0.52 & 0.54 & 0.53 \\
                  &                 &      &      &    \\
bert              & 0.61          & 0.59 & 0.64 & 0.61 \\
sngram            & 0.53          & 0.53 & 0.53 & 0.53 \\
psych             & 0.52          & 0.51 & 0.52 & 0.52 \\
bert+sngram       & \textbf{0.62} & 0.62 & 0.62 & 0.62 \\
bert+psych        & 0.60          & 0.60 & 0.60 & 0.60 \\
                  &                 &      &      &    \\
bert+sngram+psych & 0.59          & 0.58 & 0.59 & 0.59 \\ 
\hline
\end{tabular}
\end{table}

Once again, the best-performing model is {\M bert+sngram}, but  others, including the baseline in \cite{bertha2019}, were found to be similar. The corresponding homogeneous groups are illustrated in Table \ref{tab.t5.homo}.

\begin{table}[htb!]
\centering
\caption{Author-level political stance detection homogeneous groups. }
\label{tab.t5.homo}
\begin{tabular}{l c c c c c}
\hline
Model             & Acc & \multicolumn{3}{c}{Groups} \\ 
\hline
bert+sngram       & \textbf{0.62} & A &   \\
bert              & 0.61          & A &   \\
bert+psych        & 0.60          & A &   \\
bert+sngram+psych & 0.59          & A &   \\
Jiang et. al.     & 0.58          & A &   \\
\mbox{BERT.baseline} & 0.53          &   & B \\
sngram            & 0.53          &   & B \\
psych             & 0.52          &   & B \\ 
\hline
\end{tabular}
\end{table}

As in some of the previous experiments, although {\M bert+sngram} still obtains the highest accuracy among the alternatives, some of its simpler sub-components (as  \textit{bert}, in this case) were found to be statistically similar. Moreover, the full model {\M bert+sngram+psych} is once again statistically similar to the reference baseline in \cite{bertha2019}. 



\section{Discussion}
\label{sec.discuss}

In what follows we summarise our main results (Section \ref{sec.summary}) and report prediction explanations for the main classification tasks.

\subsection{Results summary}
\label{sec.summary}

Table \ref{tab.clusters} shows the tasks in which each of the  models  under evaluation ranks among the top-performing alternatives (i.e., belonging to cluster A in each of the homogeneous groups illustrated in the previous section.)

\begin{table}[htb!]
\centering
\caption{Models that rank in the top-performing cluster (A) for each task.}
\label{tab.clusters}
\begin{tabular}{l c | ccc | cccc}
\hline
\multicolumn{2}{c|}{} &
\multicolumn{3}{c|}{Text-level tasks} & 
\multicolumn{4}{c}{Author-level tasks} \\ 
Model               &Wins   & T1  & T2.bin & T2.ter & T3 & T4.bin & T4.ter & T5\\
\hline
Jiang et al.        &5&     &     A   &    A    &     &   A     &     A   & A  \\
\mbox{BERT.baseline}&1&     &         &         &     &   A     &         &    \\
                    & &     &         &         &     &         &         &    \\
bert                &6&  A  &     A   &    A    &  A  &         &     A   &  A \\
sngram              &1&     &         &         &     &         &     A   &    \\
psych               &1&     &         &         &     &   A     &         &    \\
bert+sngram         &7&  A  &     A   &    A    &  A  &   A     &     A   &  A \\
bert+psych          &5&  A  &     A   &         &  A  &   A     &         &  A \\
                    & &     &         &         &     &         &         &    \\
bert+sngram+psych   &3&     &         &         &     &   A     &     A   &  A \\
\hline  
\end{tabular}
\end{table}

From these results, a number of observations are warranted. First, we notice that there is a large number of statistically-similar models in the author-level tasks (on the right of the table), particularly in the case of the essay (tasks T4) and Twitter (task T5) domains. This was to some extent to be expected as the information to be learned from author-level inference is less explicit in the input texts, making these tasks  possibly more challenging than text-level inference in general. 

Second, we notice that the full model, comprising the main CNN architecture and \textit{bert}, \textit{sngram} and \textit{psycho} sub-components, is generally not the best choice, being often outperformed by simpler alternatives and/or by the reference baseline system in \cite{bertha2019}. 

Regarding the role of individual sub-components of the main architecture, we notice that using \textit{psych} or \textit{sngram} alone is clearly insufficient, and that even the  use of the \textit{bert} component alone generally fails to deliver optimal results. This outcome suggests that using fine-tuned BERT and the CNN architecture alone, as in the present \textit{bert} model (not to be mistaken by the standard {\M BERT.baseline} model in the second row from the top, which does not use a CNN classifier) explains much of the best results obtained across our experiments, but not all of them. In fact,  it is the combination of BERT and \textit{sngrams} in the CNN architecture (represented by the {\M bert+sngram} model) that generally obtains the best results among these alternatives, being  the top-performing model in 5 out of 7 tasks. This outcome may be partially explained by the simultaneous use of two representations of the input text (i.e., linearly ordered tokens and count-based syntactic bigram features), but we notice that larger models using this strategy (e.g., including psycholinguistics-motivated features) are not necessarily better.

Finally, we notice that the baseline in \cite{bertha2019} remains highly competitive and, although seldom obtaining the highest accuracy, it is often found within the group of top-performing alternatives. 

\subsection{Feature importance}

As a means to illustrate the word features more strongly correlated with each class and task, we performed eli5 model explanation to obtain word weights representing the change (decrease/increase) of the evaluation score when a given feature is shuffled.

Table \ref{tab.eli5.semeval} presents the most important features associated with the hyperpartisan (i.e., non-neutral) class in the  SemEval tasks T1 ({\M by\_articles}) and T3 ({\M by\_publisher}) datasets.    
  
\begin{table}[htb!]
\centering
\caption{Text-based (left) and author-based (right) hyperpartisan news detection most important features.}
\label{tab.eli5.semeval}
\begin{tabular}{c l | c l}
\hline
\multicolumn{2}{c |}{T1: Text-level} &
\multicolumn{2}{c}{T3: Author-level}  \\ 
weight & feature & weight & feature \\
\hline
10.467 & cnn & 6.02 & window \\
10.169 & american & 5.585 & bush \\
10.069 & it & 5.542 & mr \\
9.905 & women & 5.149 & that \\
9.708 & hillary & 4.833 & iraq \\
9.536 & her & 4.367 & nyse \\
9.507 & now & 4.089 & billion \\
... & ... & ... & ... \\
-6.754 & million & -4.699 & reuters \\
-6.756 & muslim & -4.829 & percent \\
-6.952 & rally & -5.025 & globalpost \\
-7.46 & donald & -5.463 & mexico \\
-7.531 & isis & -5.629 & california \\
-7.895 & <BIAS> & -6.268 & albuquerque \\
-8.17 & mr & -6.921 & ap \\
\hline  
\end{tabular}
\end{table}

Table \ref{tab.eli5.brmoral} presents the most important features associated with the left/right classes in the BRmoral corpus tasks T2 and T4 ({\M by\_opinion} and  {\M by\_author} datasets, respectively.)

\begin{table}[htb!]
\centering
\caption{Text-level (left) and author-level (right) political orientation  detection most important features.}
\label{tab.eli5.brmoral}
\begin{tabular}{c l c l   |   c l c l}
\hline
\multicolumn{4}{c |}{T1: Text-level} &
\multicolumn{4}{c}{T3: Author-level}  \\ 
\multicolumn{2}{c }{left orientation} &
\multicolumn{2}{c|}{right orientation}  &
\multicolumn{2}{c }{left orientation} &
\multicolumn{2}{c}{right orientation} \\  
weight & feature & weight & feature & weight & feature & weight & feature\\
\hline
2.533 & remove & 2.743 & black & 0.902 & women & 0.628 & favour \\ 
2.388 & guns & 2.526 & choice & 0.805 & about & 0.618 & mine \\ 
2.385 & institution & 2.493 & love & 0.739 & lay & 0.618 & be (present) \\ 
2.29 & am & 2.278 & marriage & 0.668 & society & 0.6 & since \\ 
2.234 & churches & 2.195 & wed & 0.654 & more & 0.587 & acts \\ 
1.86 & money & 2.174 & couple & 0.643 & body & 0.549 & should \\ 
1.807 & god & 2.154 & abortion & 0.625 & population & 0.54 & be (subjunctive) \\ 
1.751 & youth & 1.977 & alcohol & 0.603 & while & 0.533 & freedom \\ 
1.727 & life & 1.965 & traffic & 0.596 & alcohol & 0.481 & they \\ 
1.696 & for & 1.944 & gender & 0.595 & think & 0.465 & I \\ 
... & ... & ... & ... & ... & ... & ... & ... \\ 
-1.677 & black & -1.738 & reduce & -0.515 & defend & -0.483 & she \\ 
-1.698 & child & -1.746 & innocent & -0.515 & mine & -0.494 & country \\ 
-1.851 & law & -1.817 & church & -0.516 & they & -0.494 & are \\ 
-1.857 & crimes & -1.953 & remove & -0.528 & crimes & -0.507 & population \\ 
-1.984 & favour & -2.111 & although & -0.546 & because & -0.523 & this\_way \\ 
-2.017 & consciousness & -2.114 & have & -0.572 & same & -0.526 & women \\ 
-2.029 & alcohol & -2.138 & institution & -0.59 & hideous & -0.551 & some \\ 
-2.083 & hideous & -2.394 & churches & -0.633 & colour & -0.621 & or \\ 
-2.285 & prohibition & -2.505 & guns & -0.703 & am & -0.621 & public \\ 
-2.416 & freedom & -3.19 & perhaps & -0.755 & be & -0.658 & lay \\ 
\hline  
\end{tabular}
\end{table}

Finally, Table \ref{tab.eli5.govbr} presents the most important features associated with political stance in GovBR task T5.    
  
\begin{table}[htb!]
\centering
\caption{Political stance detection most important features.}
\label{tab.eli5.govbr}
\begin{tabular}{c l | c l}
\hline
weight & feature & weight & feature \\
\hline
5.826 & supreme\_court  & -4.087 & came \\
5.709 & jair\_bolsonaro  & -4.138 & let's\_go \\
5.159 & communists  & -4.161 & some \\
4.875 & visit  & -4.201 & poor \\
4.789 & senator  & -4.235 & human \\
4.694 & criminal  & -4.33 & (president, derogatory) \\
4.653 & red  & -4.399 & out \\
4.589 & care  & -4.662 & sportive \\
4.561 & call  & -4.687 & damn \\
4.507 & thinks  & -4.899 & political \\
\hline  
\end{tabular}
\end{table}

\section{Final remarks}
\label{sec.final}

This paper has addressed the issue of  how to combine  transformed-based text representations, which are main stream in NLP and related fields,  with  both syntactic dependency information  and psycholinguistics-motivated features for political inference from text data. In doing so, we considered both text- and author-level task definitions in both English and Portuguese languages, and introduced a novel dataset devoted the latter. 

As expected in experiments involving a range of tasks, datasets and languages, our present results vary considerably across evaluation settings and, although BERT remains a robust baseline for many of these tasks, in most cases it was possible to obtain significant improvements by making use of additional representations. This is not to say, however, that combining \textit{all} current three text representations (BERT, syntax and psycholinguistics) into a single model is the best option. In particular, it was a subset of our original CNN architecture combining only BERT and the syntactic dependency model that obtained overall best results in most tasks. 

The current work leaves a number of opportunities for further research. First, we notice that political bias and ideology are relatively broad terms that may actually include a wide range of distinct politically-related phenomena, and that future NLP studies may benefit from more fine grained task definitions. 

We notice also that many other pre-trained language models have been made available in recent years, including ELMo \cite{elmo}, XLNet \cite{xlnet}, RoBERTa \cite{roberta}, GPT-3 \cite{gpt3}. Whether any of these may outperform BERT when combined with other text representations (as in the present work) remains an open research question. 

Finally, the present use of text representations is only a first step towards more informed models that may ultimately combine BERT with many other text- and author-level features. In particular, the present architecture may be expanded with, for instance, sentiment or emotion-related information, or even with author demographics (e.g., gender, age, personality traits, moral foundations etc.)  obtained with the aid of  author profiling classifiers \cite{ca-fineg}. These and many other related possibilities are also left as future work.


\subsection*{Author contributions}
{\bf Samuel Caetano da Silva}: Conceptualisation, Methodology, Software, Validation, Formal analysis, Investigation,  Writing - original draft, reviewing and editing. {\bf Ivandr\'e Paraboni}: Conceptualisation, Writing - review and editing, supervision.

\subsection*{Financial disclosure}
The first author received support from the Brazilian Foundation CAPES - Coordination for the Improvement of  Higher Education Personnel, under grant  88882.378103/2019-01.

\subsection*{Conflict of interest}
The authors declare no potential conflict of interests.

\bibliography{refs}%

\clearpage  

\section*{Author Biography}

\begin{biography}

{\includegraphics[height=66pt]{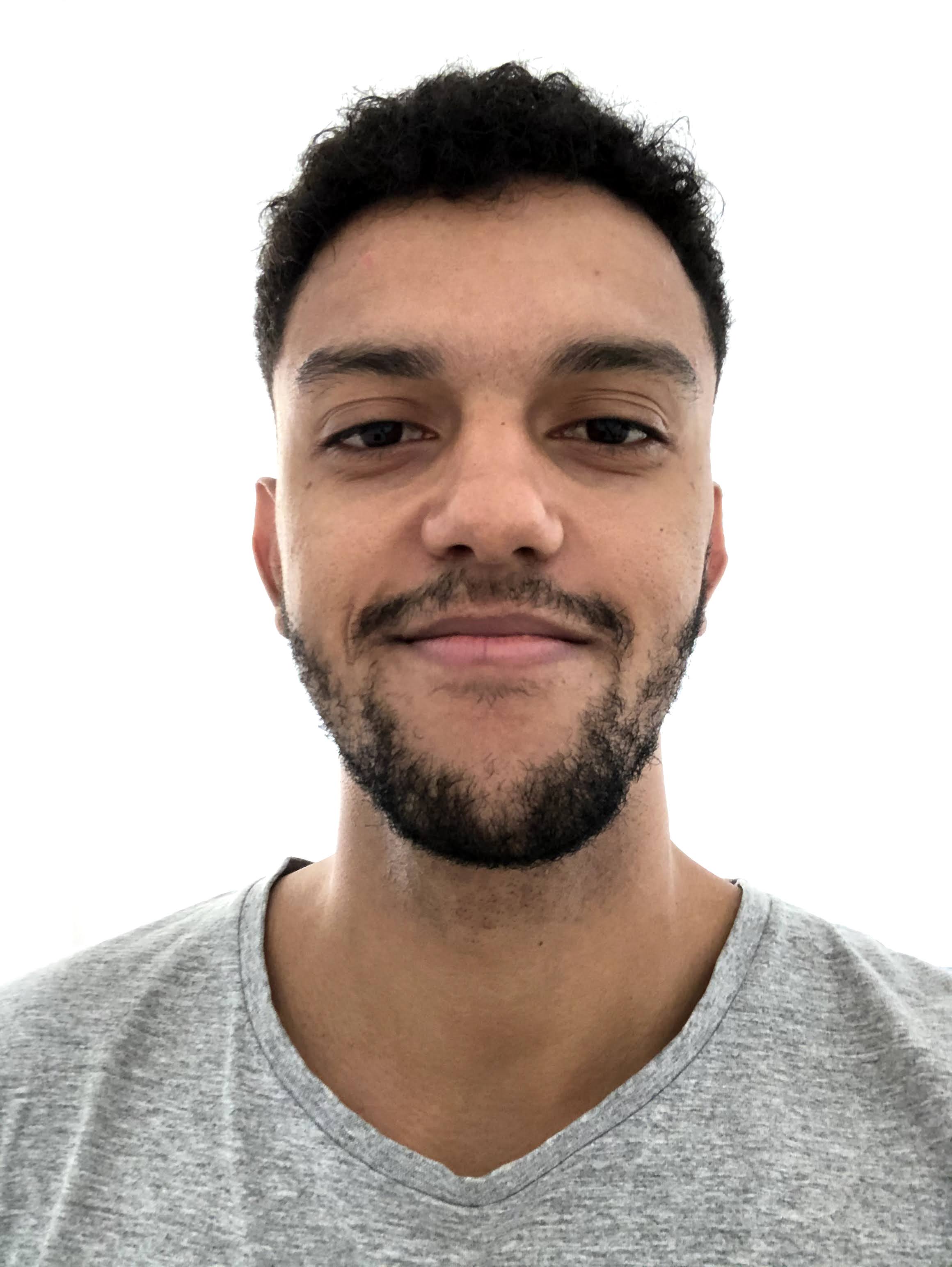}}\noindent
{\textbf{Samuel Caetano da Silva.} Graduate student of Information Systems at the University of S\~ao Paulo.\\[4ex]}

{\includegraphics[height=66pt]{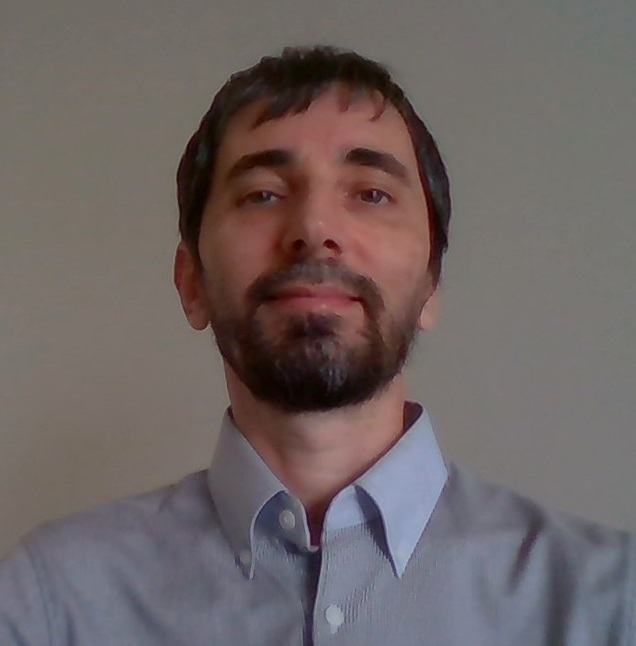}}{\textbf{\\Ivandr\'e Paraboni.} PhD in Computer Science (Univ. of Brighton, UK, 2003), and associate professor at the University of S\~ao Paulo.}\\[1ex]

\end{biography}

\end{document}